\documentclass[runningheads]{llncs}

\usepackage[T1]{fontenc}
 \usepackage{amssymb}
 \usepackage{amsmath}
 \usepackage{subcaption}
\usepackage{graphicx,verbatim}
\usepackage{xcolor}

\begin{document}
\title{AbdomenGen: Sequential Volume-Conditioned Diffusion Framework for Abdominal Anatomy Generation}
\titlerunning{AbdomenGen}
%

\author{Yubraj Bhandari$^{1,3}$, Lavsen Dahal$^{1,2}$, Paul Segars$^{1}$, Joseph Y. Lo$^{1,2}$}  
\authorrunning{Anonymized Author et al.}
\institute{$^1$Center for Virtual Imaging Trials, Dept. of Radiology, Duke University School of Medicine \\ 
$^2$ Dept. of Electrical and Computer Engineering, Pratt School of Engineering, Duke University \\
$^3$ Dept. of Physics, Trinity School of Arts and Sciences, Duke University}
  
\maketitle              
\begin{abstract}
Computational phantoms are widely used in medical imaging research, yet current systems to generate controlled, clinically meaningful anatomical variations remain limited. We present AbdomenGen, a sequential volume-conditioned diffusion framework for controllable abdominal anatomy generation. We introduce the \textbf{Volume Control Scalar (VCS)}, a standardized residual that decouples organ size from body habitus, enabling interpretable volume modulation. Organ masks are synthesized sequentially, conditioning on the body mask and previously generated structures to preserve global anatomical coherence while supporting independent, multi-organ control. Across 11 abdominal organs, the proposed framework achieves strong geometric fidelity (e.g., liver dice $0.83 \pm 0.05$), stable single-organ calibration over $[-3,+3]$ VCS, and disentangled multi-organ modulation. To showcase clinical utility with a hepatomegaly cohort selected from MERLIN, Wasserstein-based VCS selection reduces distributional distance of training data by 73.6\% . These results demonstrate calibrated, distribution-aware anatomical generation suitable for controllable abdominal phantom construction and simulation studies.
\end{abstract}

\section{Introduction}

Computational anatomical phantoms underpin virtual imaging trials, radiation dosimetry, and surgical simulation~\cite{segars2010xcat,xu2014computationalphantoms,chheang2021collaborative}. As deep learning systems are deployed, retrospective data still undersample rare, extreme, and pathological anatomies, precisely where failures concentrate. Phantom populations enable controlled stress-testing by generating and enriching such regimes on demand~\cite{guo2024maisi}. A useful population therefore requires fidelity (realistic, spatially coherent organs), diversity (habitus/morphology/inter-organ variation), and targeted controllability (clinically grounded, calibrated enrichment).

To our knowledge, no existing method satisfies all three requirements for multi-organ abdominal shape synthesis with habitus-aware, organ-wise control. Model-based phantom frameworks, such as the XCAT and related extensions ~\cite{segars2010xcat,dahal2025xcat}, range from manual construction to scalable deep learning pipelines, but do not support deliberate or calibrated anatomical manipulation. Fully generative approaches offer more flexibility, yet remain limited to single-organ synthesis or uncalibrated latent interpolation~\cite{li2023anatomycompletor,podobnik2025anatomygen,mouheb2025largeintestine}. MAISI targets paired CT image and mask synthesis with coarse volume conditioning, but shape controllability is secondary and does not support independent organ-wise manipulation within a fixed anatomical context~\cite{guo2024maisi}. Habitus-calibrated, organ-wise controllable multi-organ shape synthesis remains unaddressed.

We propose \textbf{AbdomenGen}, a sequential, volume-conditioned diffusion framework for controllable abdominal phantom generation over 11 structures.The framework enables the generation of anatomically coherent phantom populations with calibrated, organ-specific variability for use in medical imaging research. We target the abdomen because its combination of deformable solid organs, filling-dependent hollow structures, and compact organs with tight spatial margins imposes strong inter-organ geometric constraints that stress every aspect of controllable generation. Two core challenges motivate our design: conditioning on raw organ volume produces contradictory signals across body habitus, and joint multi-organ generation struggles to maintain spatial consistency. We address the first with the \textbf{Volume Control Scalar (VCS)}, a habitus-decoupled residual that provides a interpretable control axis, and the second with a sequential context-aware strategy in which each organ is synthesized conditioned on the body mask and all previously generated structures.

\noindent\textbf{Contributions:}
\begin{enumerate}
\item \textbf{AbdomenGen}: a scalable diffusion framework that generates 11 abdominal organs sequentially one at a time, each conditioned on the body mask and all prior structures, achieving spatial coherence and independent organ-wise control across the full abdominal cavity.
\item \textbf{Volume Control Scalar (VCS)}: a habitus-decoupled conditioning variable that resolves the contradictory-signal problem of raw-volume conditioning, providing a calibrated and interpretable control axis to modulate organ size while remaining grounded in body-scale statistics.
\item \textbf{Distribution-level controllability}: By assigning VCS distributions, we can match a target patient population, as shown by the Wasserstein-optimal reduction of distributional distance for a hepatomegaly population by 73.6\% without retraining, demonstrating on-demand coverage of clinically meaningful anatomical regimes.
\end{enumerate}

\section{Method}

\begin{figure}[!t]
    \centering
    \includegraphics[width=1.0\linewidth]{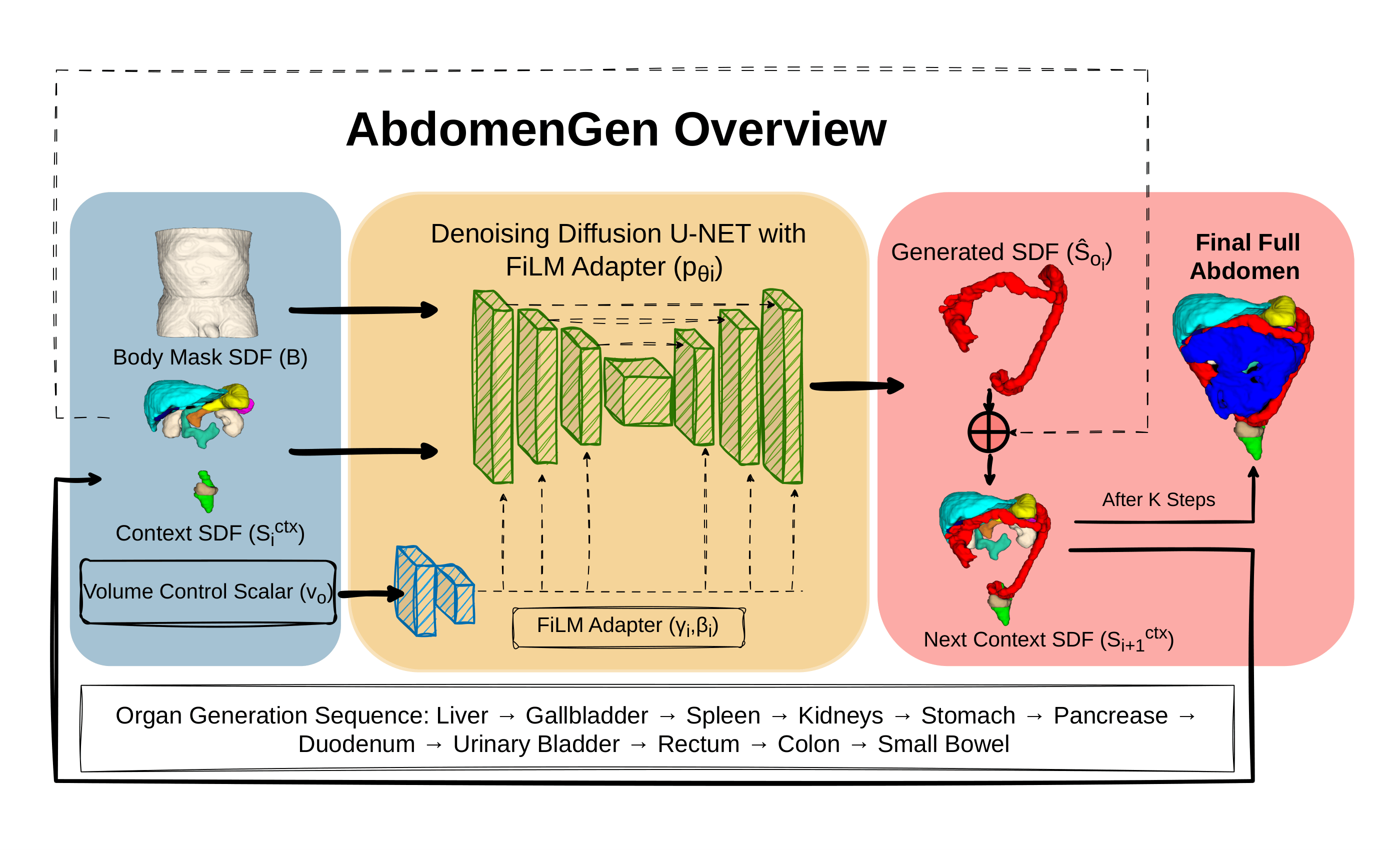}
    \caption{
    Overview of AbdomenGen, sequential organ generation framework based on denoising diffusion U-Net with FiLM conditioning.
    For generation of organ $o_i$ (illustrated left to right for colon), the model $p_{\theta_i}$ takes as input the body mask signed distance field (SDF) $B$, current context SDF $S_i^{\mathrm{ctx}}$, and volume control scalar $v_o$.
    The scalar is injected through a FiLM adapter to modulate intermediate network features. The generated organ SDF $\hat{S}_{o_i}$ is obtained and incorporated into the context to form $S_{i+1}^{\mathrm{ctx}}$. Organs are generated in a predefined order (bottom), and the accumulated outputs form the final full abdominal anatomy.
}
    \label{fig:method_overview}
    
\end{figure}

\textbf{Problem Formulation.}
Organ synthesis is modeled as conditional generation of an organ signed distance field (SDF). 
For organ $o$, let $S_o$ denote its SDF, $B$ the body-mask SDF, $S^{\mathrm{ctx}}_o$ an aggregated context SDF from surrounding organs, and $v_o\in\mathbb{R}$ a scalar size control. We learn
\[
p_{\theta_o}\!\left(S_o \mid B, S^{\mathrm{ctx}}_o, v_o\right).
\]
The shapes are represented as SDFs following prior implicit and diffusion-based modeling~\cite{takikawa2021neural,chou2023diffusion}, which provides a smooth geometric signal and supports training / inference on a fixed grid.



\textbf{Sequential Generation Strategy. }
We synthesize the abdomen by generating organs sequentially in a fixed order, from large spatially defining structures to smaller, more dependent organs. For organ $o_i$, a fixed-size context SDF by aggregating previously generated organs is formed as:
$
S_i^{\mathrm{ctx}} = \max_{j<i} S_{o_j}.
$
This representation encodes occupied space without increasing input channels as $i$ grows. During training, $S^{\mathrm{ctx}}_i$ is built from reference mask; at inference it is constructed autoregressively from generated predictions.

\textbf{Volume Control Scalar (VCS).}
We aim to explicitly control organ size, but organ volume is coupled to global body habitus. Prior work reports associations between liver/spleen volume and anthropometric measures~\cite{livervolumeolthof2019accuracy,prassopoulos1997determinationspleenvol}, and we observe similar trends: organ volume is partially predictable from body mask volume, with Pearson correlations of $r=0.64$ (liver), $r=0.52$ (kidneys), $r=0.24$ (spleen), and $r=0.28$ (stomach). Conditioning on raw organ volume would therefore partially re-encode information already provided by the body mask $B$ and may request body--organ volume pairs not supported by the training distribution.

To decouple size control from habitus, we model the expected organ volume as a linear function of body volume,
\[
\hat{V}_{o_i}(V_B)=a_i V_B + b_i,
\]
with $(a_i,b_i)$ fit on the training set. We define the residual and its standardized form (VCS) as
\begin{equation}
r_{o_i}=V^{\mathrm{ref}}_{o_i}-\hat{V}_{o_i}(V_B), \qquad
v_i=\frac{r_{o_i}-\mu_i}{\sigma_i},
\label{eqn:1}
\end{equation}
where $(\mu_i,\sigma_i)$ are the mean and standard deviation of residuals for organ $i$. We apply this uniformly across organs for consistency; even when correlations are weak (e.g., gallbladder), residualization keeps the control variable decorrelated from global body scale while preserving anatomical coherence.

\textbf{Diffusion Architecture with FiLM.}
A denoising diffusion model is trained per organ~\cite{ho2020denoising}. At timestep $t$, the network input is
$
x_t = [S^{\mathrm{noisy}}_{o,t};\, B;\, S^{\mathrm{ctx}}_o],
$
i.e., the noisy organ SDF, the body-mask SDF $B$, and a context SDF encoding surrounding organs. We condition on the volume control scalar $v_o$ using FiLM~\cite{perez2018film} and use $x_0$-parameterization to predict the clean SDF:
$
\widehat{S}_o = f_{\theta_o}(x_t, t, v_o).
$



\textbf{Training Objective.}
Let $\widehat{S}_{o_i}$ denote the predicted SDF, $S^{\mathrm{ref}}_{o_i}$ the reference mask, and soft occupancy map is obtained via sigmoid function as $\widehat{M}_{o_i} = \sigma(10 \odot \widehat{S}_{o_i})$ and $M^{\mathrm{ref}}_{o_i}$ is obtained by simple thresholding.  
We use an $\ell_1$ reconstruction loss $\mathcal{L}_{\mathrm{SDF}}$ and binary cross-entropy loss $\mathcal{L}_{\mathrm{BCE}}$ as,
\[
\mathcal{L}_{\mathrm{SDF}} = \|\widehat{S}_{o_i} - S^{\mathrm{ref}}_{o_i}\|_1
\qquad
\mathcal{L}_{\mathrm{BCE}}
= \mathrm{BCE}\!\left(\widehat{M}_{o_i}, M^{\mathrm{ref}}_{o_i}\right).
\]

Additionally, we introduce a differentiable reverse Dice penalty $\mathcal{L}_{\mathrm{ov}}$ to discourage context-organ overlap. 
After the initial training epochs, we activate a VCS calibration loss $\mathcal{L}_{\mathrm{vcs}}$, where the predicted organ volume (computed from the generated mask) yields a standardized residual $\widehat{v}_{i}$ that is matched to the target control $v_{i}$:
\[
\mathcal{L}_{\mathrm{ov}} =
1 -
\frac{
2 \sum_{x} \hat{M}_i(x)\, M_i^{\mathrm{ctx}}(x)
}{
\sum_{x} \hat{M}_i(x)
+
\sum_{x} M_i^{\mathrm{ctx}}(x)
+
\epsilon
},
\qquad
\mathcal{L}_{\mathrm{vcs}}
=
\left\| \hat{v}_i - v_i \right\|_2^2.
\]

Thus, the total loss function is,
$
\mathcal{L} = \mathcal{L}_{\mathrm{SDF}} 
+ \mathcal{L}_{\mathrm{BCE}} 
+ \mathcal{L}_{\mathrm{ov}} 
+ \mathcal{L}_{\mathrm{vcs}}
$

\section{Experimental Setup}

\noindent\textbf{Data.}
We use 556 abdominal CT scans from a [redacted] U.S. academic hospital, split into 314 training, 42 internal validation, and 200 held-out test cases. Multi-organ masks were produced by the organ segmentation framework DukeSeg~\cite{dahal2025xcat} and reviewed by a physician for quality control. These masks serve as reference anatomical supervision for training and evaluation. We also evaluate distribution matching on an external hepatomegaly cohort (from MERLIN)~\cite{blankemeier2024merlin}. Volumes are cropped to the abdominal region (liver to rectum), resampled to a $128^3$ grid, and converted to signed distance functions.

\noindent\textbf{Training and inference.}
We implement a 3D diffusion U-Net in MONAI~\cite{cardoso2022monai} with channel widths $[32,64,64,128,256]$. Models are trained for 1000 linear-$\beta$ diffusion steps to predict $x_0$ (clean SDF) using AdamW ($10^{-4}$ learning rate, $10^{-4}$ weight decay). For robustness to missing controls, we drop conditioning signals with probability $p=0.3$. At inference, we sample using 10-step DDIM~\cite{song2020denoising}.

\noindent\textbf{Evaluation.}
After rigid ANTs registration~\cite{tustison2021antsx}, we report Dice, and extract surface points with PCA alignment to report ASSD, HD95, and symmetric Chamfer distance (mm). Manifold realism is assessed by nearest-training-neighbor distances (Chamfer and HD95) for both generated and test samples. Diversity is measured via pairwise Dice/Chamfer dispersion among generated livers, and distribution matching is quantified by the 1D Wasserstein-1 distance ($W_1$) between generated and target liver-volume distributions across VCS sweeps.

\section{Results}
We report (i) shape fidelity and manifold realism, (ii) calibrated single-organ volume control, (iii) multi-organ independence under joint conditioning, and (iv) distribution-level matching to a hepatomegaly cohort.

\begin{table*}[t]
\centering
\setlength{\tabcolsep}{5pt}
\renewcommand{\arraystretch}{1.1}
\fontsize{8}{12}\selectfont
\caption{
(A) Geometric agreement between generated masks and reference mask (Gen vs Ref). 
(B) Manifold realism: distance from generated masks to their nearest training sample (Gen vs Train-NN). 
(C) Reference training variability: distance from reference masks to their nearest training sample (Ref vs Train-NN). 
Values are reported as mean $\pm$ std across 200 test cases.
}
\label{tab:shape_fidelity}
\begin{tabular}{lcccc}
\hline
& \textbf{Liver} & \textbf{Spleen} & \textbf{Kidneys} & \textbf{Stomach} \\
\hline
\multicolumn{5}{l}{\textbf{(A) Fidelity: Generated vs Reference Masks}} \\
Dice $\uparrow$ 
& 0.83 $\pm$ 0.05 
& 0.68 $\pm$ 0.13 
& 0.60 $\pm$ 0.23 
& 0.68 $\pm$ 0.15 \\

ASSD (mm) $\downarrow$ 
& 5.85 $\pm$ 2.34 
& 6.23 $\pm$ 2.83 
& 6.16 $\pm$ 6.72 
& 6.36 $\pm$ 4.16 \\

HD95 (mm) $\downarrow$ 
& 17.58 $\pm$ 9.35 
& 18.61 $\pm$ 9.05 
& 16.05 $\pm$ 22.50 
& 21.26 $\pm$ 17.76 \\

Chamfer (mm) $\downarrow$ 
& 11.70 $\pm$ 4.68 
& 12.47 $\pm$ 5.67 
& 12.33 $\pm$ 13.44 
& 12.71 $\pm$ 8.33 \\
\hline

\multicolumn{5}{l}{\textbf{(B) Manifold realism: Generated vs Train-Nearest Neighbor}} \\
HD95 (mm) $\downarrow$ 
& 13.92 $\pm$ 2.78 
& 7.66 $\pm$ 2.20 
& 10.49 $\pm$ 19.17 
& 10.09 $\pm$ 2.50 \\

Chamfer (mm) $\downarrow$ 
& 11.20 $\pm$ 1.97 
& 6.91 $\pm$ 1.64 
& 8.47 $\pm$ 11.67 
& 8.56 $\pm$ 1.57 \\
\hline

\multicolumn{5}{l}{\textbf{(C) Reference: Reference Masks vs Train-Nearest Neighbor}} \\
HD95 (mm) $\downarrow$ 
& 14.87 $\pm$ 3.46 
& 8.09 $\pm$ 2.45 
& 6.02 $\pm$ 0.96 
& 10.51 $\pm$ 2.66 \\

Chamfer (mm) $\downarrow$ 
& 12.04 $\pm$ 2.41 
& 7.19 $\pm$ 1.68 
& 5.82 $\pm$ 0.74 
& 8.95 $\pm$ 1.93 \\
\hline
\end{tabular}

\end{table*}

\subsection{Shape Fidelity and Diversity}

Table~\ref{tab:shape_fidelity} evaluates volumetric overlap (Dice), average surface deviation (ASSD), worst-case boundary error (HD95), and global surface discrepancy (Chamfer). AbdomenGen achieves strongest geometric agreement for the liver, with high overlap and low boundary error, while performance decreases for smaller or more geometrically variable organs. The larger variance observed for kidneys is reflected primarily in surface-based metrics (HD95, Chamfer), indicating localized boundary deviations rather than systematic volumetric failure.

To rule out memorization, we compare each generated sample's nearest-neighbor distance to the training set (Gen--TrainNN) against the corresponding Reference Masks distance (Ref--TrainNN). Gen--TrainNN closely tracks Ref--TrainNN across the reported organs, indicating samples remain on the empirical anatomical manifold without collapsing onto memorized training instances.

\begin{figure}[!t]
    \centering
    \includegraphics[width=0.8\linewidth]{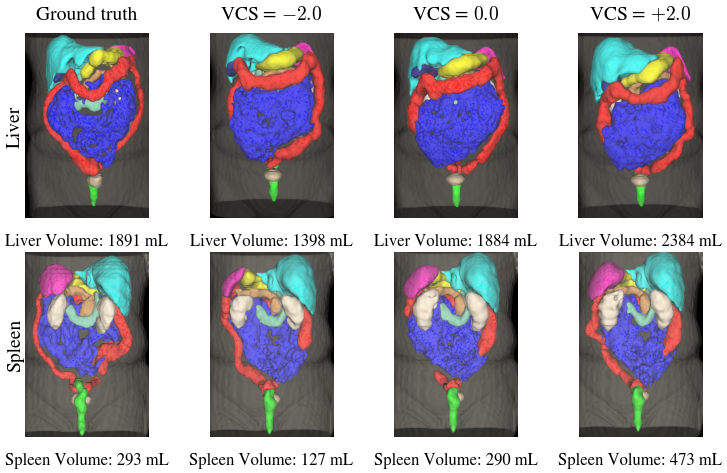}
    \caption{
\textbf{Qualitative evaluation:} Rows correspond to Liver (cyan, anteroposterior view) and Spleen (pink, posteroanterior view).  Columns (left to right) show Reference Masks and VCS $=-2$, 0, and +2 respectively. 
As VCS varies, target organ volume (below each rendering) exhibits controlled changes while surrounding anatomical structures remain spatially coherent. 
}
    \label{fig:qualitative}
\end{figure}

Figure~\ref{fig:qualitative} illustrates representative samples. Generated organs preserve topology and inter-organ spatial relationships, and varying VCS produces controlled volume changes while maintaining spatial coherence of surrounding anatomy.

\begin{table}[!t]
\centering
\setlength{\tabcolsep}{3.5pt}
\renewcommand{\arraystretch}{1.05}
\caption{\textbf{Diversity comparison between AbdomenGen and MAISI.}  Values for generated liver distributions are reported as mean $\pm$ standard deviation.}
\label{tab:liver_diversity}
\begin{tabular}{lcccccc}
\hline
\textbf{Model} 
& \textbf{Pairwise Dice}$ \downarrow $
& \textbf{Pairwise Chamfer(mm)} $\uparrow$  \\
\hline
MAISI (size parameter=0.9) 
& $0.803\pm0.057$ 
& $13.97\pm4.31$ \\

Ours (Ref Masks VCS) 
& 0.755$\pm$0.096
& 17.01$\pm$6.82 \\
\hline
\end{tabular}
\end{table}

\paragraph{Liver diversity comparison (MAISI vs. ours).}
To contextualize geometric diversity, we compare liver dispersion between MAISI (fixed size parameter = 0.9) and our volume-conditioned model (Reference Masks VCS). Table~\ref{tab:liver_diversity} shows improved inter-sample variability under AbdomenGen, reflected by lower pairwise Dice and higher pairwise Chamfer distance, with larger standard deviations consistent with patient-specific volume conditioning producing a broader range of plausible liver shapes.

\begin{figure}[!t]
    \centering
    
    \begin{subfigure}[t]{0.49\linewidth}
        \centering
        \includegraphics[width=\linewidth]{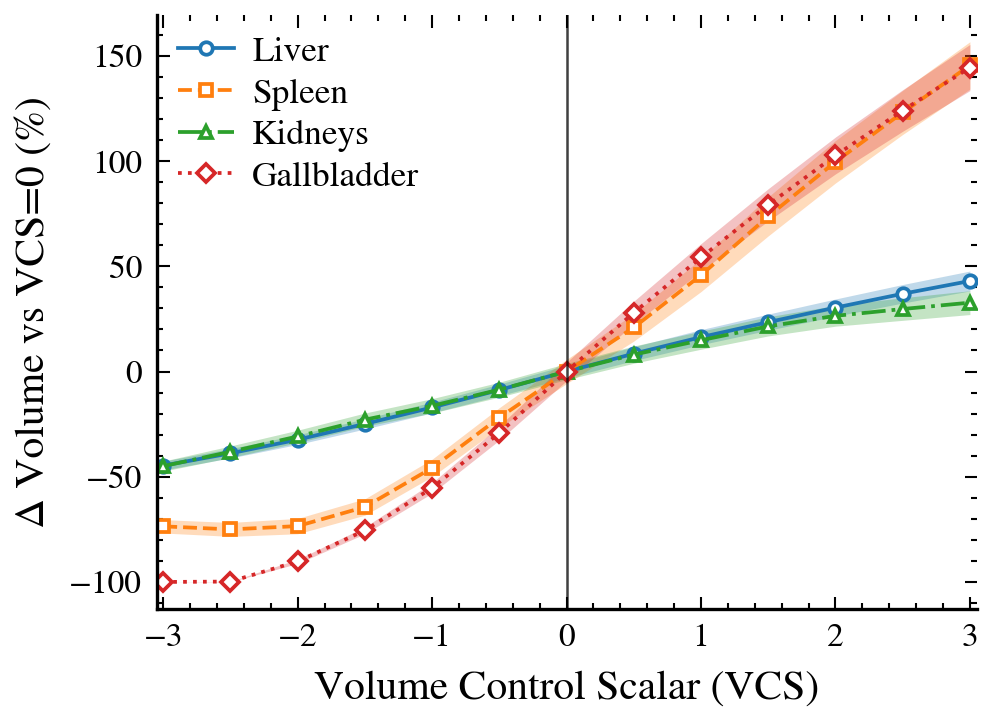}
        \caption{Cross-organ VCS calibration.}
        \label{fig:vcs_calibration}
    \end{subfigure}
    \hfill
    \begin{subfigure}[t]{0.49\linewidth}
        \centering
        \includegraphics[width=\linewidth]{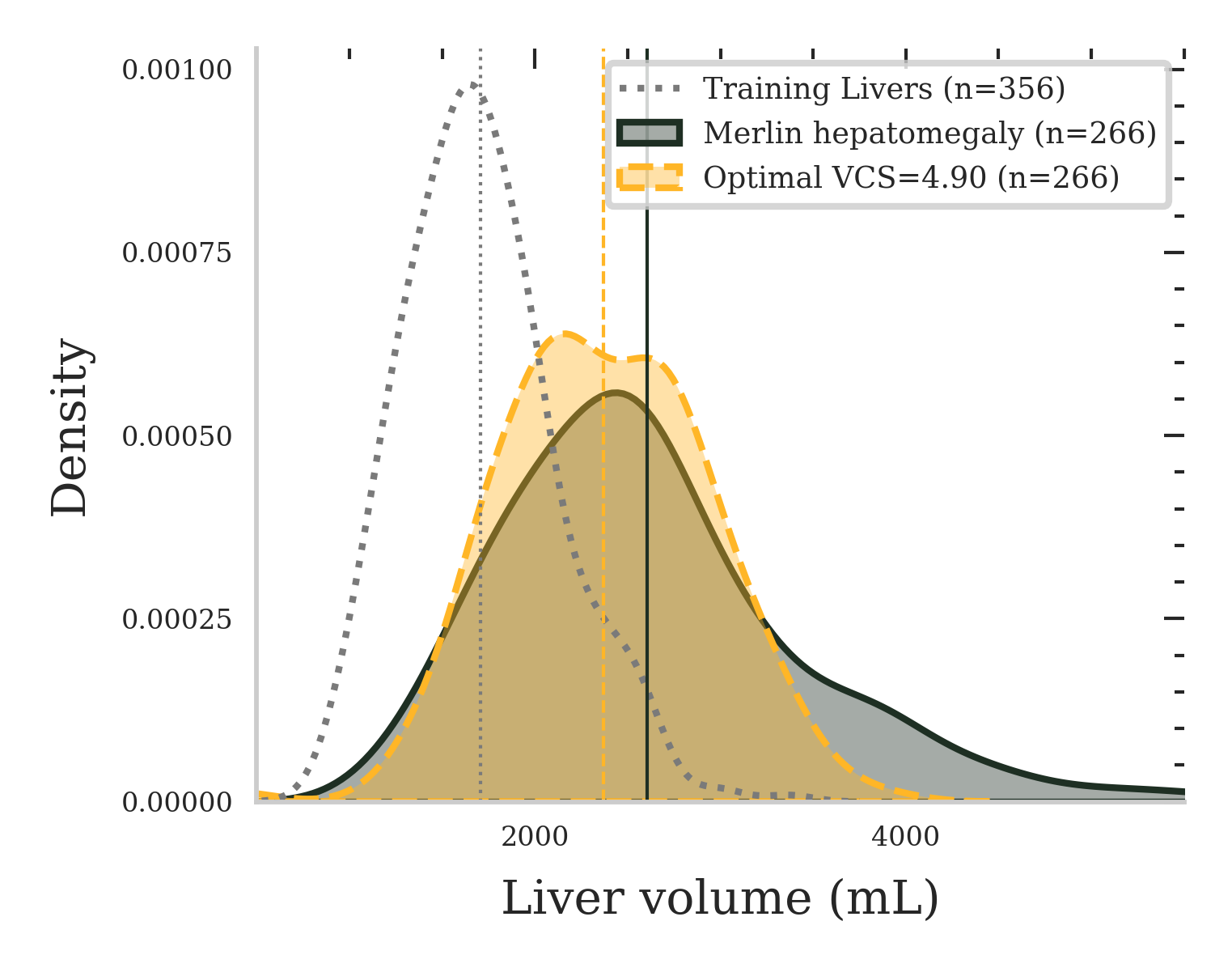}
        \caption{Distribution-level calibration.}
        \label{fig:hepatomegaly_match}
    \end{subfigure}

    \caption{
    \textbf{(a)} Mean organ volume change ($\Delta$\%) relative to VCS = 0 across 200 test cases (mean ± 95\% CI) as VCS is swept over [-3, 3]. 
    Volumes are normalized to baseline generation to enable cross-organ comparison. 
    \textbf{(b)}Kernel density estimates of liver volumes for the training cohort (dotted), Merlin hepatomegaly Reference Masks (solid), and generated samples under the selected volume-conditioning setting (dashed). Vertical dashed lines denote cohort means.
    }
    \label{fig:vcs_and_distribution}
\end{figure}

\subsection{Volume Controllability and Calibration}

\paragraph{Single-organ calibration.}
Figure~\ref{fig:vcs_and_distribution}(a) shows the mean organ volume change as VCS is swept. All evaluated organs exhibit a consistent and interpretable scaling response to VCS, indicating a stable control signal. Liver and kidneys follow an approximately linear trend, while spleen and gallbladder demonstrate amplified growth at positive VCS values. The tight 95\% confidence intervals across the sweep indicate consistent realization of the requested volume across test cases.

\paragraph{Distribution matching to hepatomegaly.}
To match a pathological target distribution, we sweep liver VCS and compute the Wasserstein-1 distance between generated liver volumes and the Merlin hepatomegaly cohort~\cite{blankemeier2024merlin}. The optimum occurs at VCS$=4.9$, reducing W$_1$ from 899.5\,mL (training vs.\ Merlin) to 237.0\,mL (73.6\%). Figure~\ref{fig:vcs_and_distribution}(b) shows that, at this setting, generated volumes closely align with the Merlin distribution while remaining distinct from the training cohort. Geometric fidelity and manifold realism remain comparable to baseline (Table~\ref{tab:shape_fidelity}).

\begin{figure}[!t]
    \centering
    \includegraphics[width=1.0\linewidth]{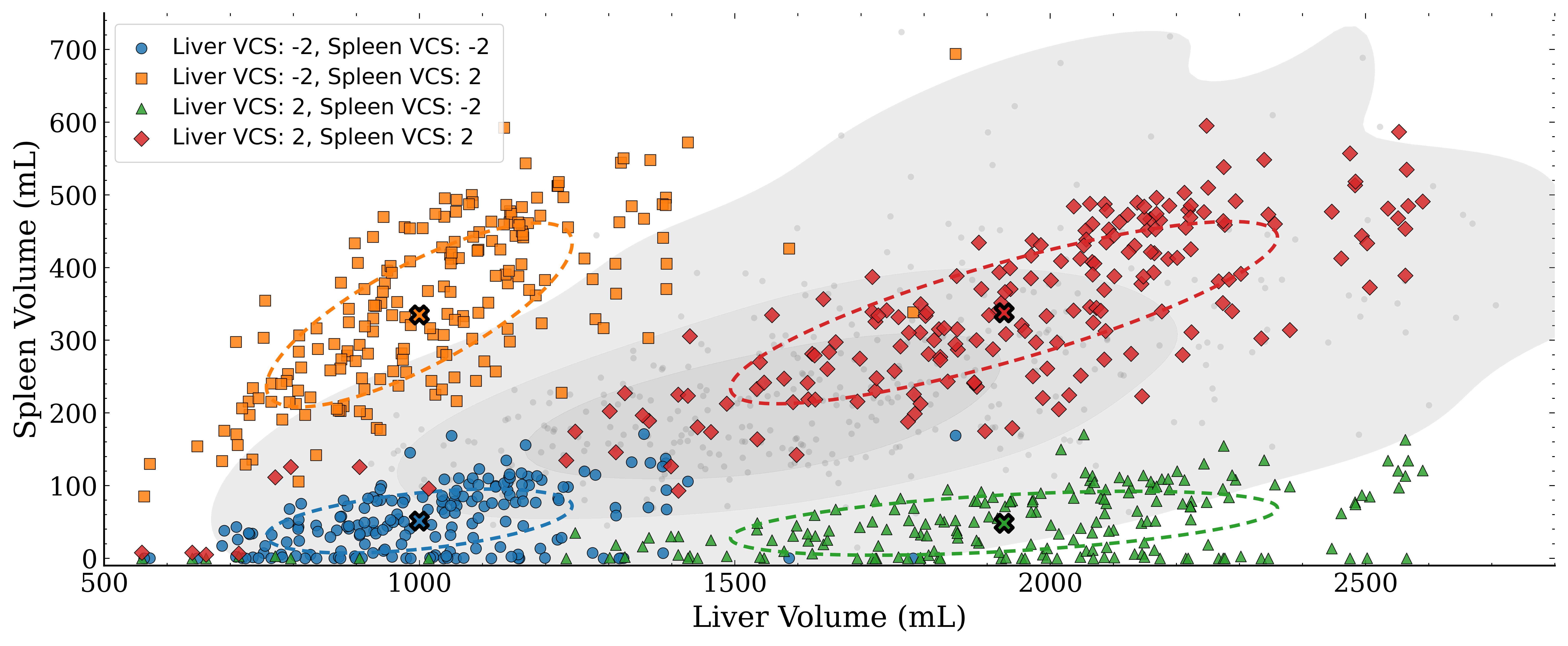}
    \caption{Joint liver–spleen volume distributions under independent VCS conditioning (±2). Gray contours denote the empirical training distribution; colored clusters show diverse but controlled generations that sampling these two dimensions.}
    \label{fig:joint_control}
\end{figure}

\paragraph{Independent multi-organ control.}
Figure~\ref{fig:joint_control} evaluates joint liver--spleen conditioning. Changing one organ's VCS shifts samples primarily along the intended axis with minimal movement in the orthogonal dimension, indicating limited cross-organ coupling. This enables targeted exploration of underrepresented joint configurations beyond high-density regions of the training distribution.

\section{Conclusion}

We introduce a statistically grounded volume-conditioning strategy for controllable anatomical generation. Our Volume Control Scalar (VCS) parameterizes organ size as a standardized residual relative to body habitus, yielding an interpretable control axis that is decorrelated from global scale. This enables calibrated, interpretable volume modulation with approximately linear response and supports stable independent control across multiple organs. By selecting VCS operating points to match clinically defined target distributions, the framework can synthesize cohort-calibrated phantom populations without latent-space interpolation heuristics or post-hoc registration.

\noindent\textbf{Limitations and future work.}
Sequential synthesis can accumulate small geometric errors across organs, and volume control alone does not fully specify local morphology, and the framework performs less reliably for elongated, tubular structures such as the colon. Future work will extend the approach to additional anatomical regions, incorporate richer conditioning signals (e.g., shape and tissue descriptors), and quantify downstream impact in image synthesis, segmentation stress-testing, and simulation pipelines.



\bibliographystyle{splncs04}
\bibliography{refs}

\end{document}